\documentclass[10pt]{article}
\usepackage[a4paper,
            margin=1in]{geometry}
\usepackage{parskip}
\usepackage[activate={true,nocompatibility},final,kerning=true,spacing=true,factor=1100,stretch=10,shrink=10]{microtype}
\usepackage{setspace}
\usepackage{amsmath}
\usepackage{amssymb}
\usepackage{graphicx}
\usepackage{booktabs}
\usepackage{sectsty}
\usepackage{tabularx}
\usepackage[normalem]{ulem}
\usepackage{csquotes}
\usepackage[htt]{hyphenat}
\usepackage{nameref}
\usepackage{mathabx}
\usepackage{totcount}
\usepackage{epigraph}

\usepackage[hyphens]{url}
\usepackage{hyperref}
\hypersetup{
     colorlinks=true,
     linkcolor=black,
     filecolor=black,
     citecolor = black,      
     urlcolor=black}    

\usepackage[style=authoryear-comp,
            natbib=true,
            backend=bibtex,
            giveninits=true,
            maxcitenames=2,
            maxbibnames=99,
            ]{biblatex}

\renewbibmacro*{cite:labeldate+extradate}{%
  \printtext[bibhyperref]{%
    \iffieldundef{note}
      {\printfield{year}\printfield{extradate}} % If no original year, just print the year
      {\printfield{note}% Prints origyear (from note)
       \addslash
       \printfield{year}\printfield{extradate}}}}       
\renewbibmacro*{finentry}{%
  \finentry
  \iffieldundef{note}
    {}% Do nothing if note is not defined
    {\newunit\newblock{ (Original work published \printfield{note})}}% Print note at the end
  }
\AtEveryBibitem{%
    \clearfield{month} % APA generally omits month of publication
    \clearfield{day}   % APA generally omits day of publication
    \clearfield{url}
    \clearfield{isbn}
    \clearfield{issn}
    \clearfield{note}
    \clearfield{urldate}
    \clearfield{address}
    \clearfield{editor}
    \clearlist{language} % Omit language field
}

\AtBeginBibliography{%
  \renewbibmacro*{urldate}{} % This removes the "visited on" date display
}

\DeclareFieldFormat{addendum}{\mkbibparens{#1}\nopunct}
\DeclareFieldFormat{edition}{\mkbibparens{#1}}
\DeclareFieldFormat[inbook]{title}{#1} % Redefine the title format for inbook
\DeclareFieldFormat[article]{title}{#1} % Italicize only journal title
\DeclareFieldFormat[book]{title}{\mkbibemph{#1}} % Italicize book titles
\DeclareFieldFormat[book]{volume}{(Vol. #1)} % Set Vol. properly
\DeclareFieldFormat{journaltitle}{\mkbibemph{#1}} % Remove quotation marks from journal titles
\DeclareFieldFormat[inproceedings]{title}{#1} % Don't italicize titles of conference papers
\DeclareFieldFormat[book]{edition}{(#1). } % Period after book edition.
\DeclareFieldFormat{pages}{#1} % no pp in page definition
\renewbibmacro*{title}{%
  \printfield{title}%
  \iffieldundef{edition}{}{\setunit{\addspace}\nopunct}%
}

\renewbibmacro{in:}{}
\renewbibmacro*{in:}{%
  \ifboolexpr{
    test {\ifentrytype{inproceedings}} % Check if the entry type is 'inproceedings'
  }
  {\bibstring{in}\intitlepunct}  % If true, print "In "
  {}            % If false, print nothing
}
\renewcommand*{\intitlepunct}{\nopunct}

\renewbibmacro*{volume+number+eid}{
  \addcomma
  \textit{\printfield{volume}}%
  \iffieldundef{number}
    {}
    {\mkbibparens{\printfield{number}}}%
  \setunit{\addcomma\space}%
  \printfield{eid}
}
\usepackage[T1]{fontenc} % font
\usepackage{mlmodern}
\usepackage[hang,flushmargin]{footmisc} 

\usepackage{etoolbox}
\AtBeginEnvironment{table}{\small}
\AtBeginEnvironment{tabular}{\small}
\AtBeginEnvironment{tabularx}{\small}
\AtBeginEnvironment{longtable}{\small}

\makeatletter
\renewcommand{\paragraph}{%
  \@startsection{paragraph}{4}{\z@}%
    {3.25ex \@plus1ex \@minus.2ex}%
    {-1em}%
    {\normalfont\fontsize{10}{12}\fontfamily{qhv}\selectfont\bfseries\protect\raisebox{-.225pt}}}
\makeatother

\makeatletter
\renewcommand{\maketitle}{
  \begin{flushleft}
    \fontfamily{qhv}\selectfont
    {\LARGE \textbf{\@title}} \\[2mm]
    \fontfamily{qhv}\selectfont \@author \\[2mm]
    \@date
  \end{flushleft}
}
\makeatother

\usepackage{caption}
\usepackage{subcaption}
\DeclareCaptionFormat{myformat}{#1#2#3\dotfill}
\newcommand{\githubrepo}{\url{https://github.com/pournaki/soteu-narratives}}

\usepackage{caption}
\usepackage{subcaption}
\DeclareCaptionFormat{myformat}{#1#2#3\dotfill}
\usepackage{graphicx}
\graphicspath{ {./figures/} }
\usepackage{booktabs}
\usepackage{tabularx}
\usepackage{url}

\title{Extracting narrative signals from public discourse:\\[2mm] a network-based approach}
\author{\vspace{1mm}\raggedright Armin Pournaki\textsuperscript{1,2,3}, Tom Willaert\textsuperscript{4,5}}%, Author 2$^{4,5}$}
\date{%
    \small
    \textsuperscript{1}Max Planck Institute for Mathematics in the Sciences, Leipzig, Germany\\%
    \textsuperscript{2}Laboratoire Lattice, École Normale Supérieure - PSL - CNRS - Univ. Sorbonne Nouvelle, Montrouge, France\\%
    \textsuperscript{3}médialab, Sciences Po, Paris, France\\%
    \textsuperscript{4}Brussels School of Governance, Vrije Universiteit Brussel, Brussels, Belgium\\%
    \textsuperscript{5}imec-SMIT, Vrije Universiteit Brussel, Brussels, Belgium\\%    
    [1ex]
     September 18, 2025
}

\begin{document}
\allsectionsfont{\usefont{T1}{qhv}{b}{n}\selectfont}

\maketitle

\vspace{7mm}
\noindent
{\fontfamily{qhv}\selectfont \large \textbf{Abstract}}
{\small
% \fontfamily{qhv}\selectfont \small  
\begin{spacing}{1.15}
Narratives are key interpretative devices by which humans make sense of political reality. As the significance of narratives for understanding current societal issues such as polarization and misinformation becomes increasingly evident, there is a growing demand for methods that support their empirical analysis. To this end, we propose a graph-based formalism and machine-guided method for extracting, representing, and analyzing selected narrative signals from digital textual corpora, based on Abstract Meaning Representation (AMR). The formalism and method introduced here specifically cater to the study of \textit{political} narratives that figure in texts from digital media such as archived political speeches, social media posts, transcripts of parliamentary debates, and political manifestos on party websites. We conceptualize these political narratives as a type of ontological narratives: stories by which actors position themselves as political beings, and which are akin to political worldviews in which actors present their normative vision of the world, or aspects thereof. We approach the study of such political narratives as a problem of information retrieval: starting from a textual corpus, we first extract a graph-like representation of the meaning of each sentence in the corpus using AMR. Drawing on transferable concepts from narratology, we then apply a set of heuristics to filter these graphs for representations of 1) actors and their relationships, 2) the events in which these actors figure, and 3) traces of the perspectivization of these events. We approach these references to actors, events, and instances of perspectivization as core narrative signals that allude to larger political narratives. By systematically analyzing and re-assembling these signals into networks that guide the researcher to the relevant parts of the text, the underlying narratives can be reconstructed through a combination of distant and close reading. A case study of State of the European Union addresses (2010 -- 2023) demonstrates how the formalism can be used to inductively surface signals of political narratives from public discourse.
\end{spacing}
}
\vspace{2mm}

\section{Introduction}

Complex societal phenomena such as political polarization and misinformation challenge analyses of public discourse based on traditional markers of political orientation. Recent scholarship indeed recognizes that these developments might be better understood in terms of how diverging interpretative frameworks shape people's understanding of reality \citep{Friedman2023}. This shifting focus towards interpretation and interpretative devices recalls a broader ``narrative turn'' in the social sciences \citep{goodson2011narrative}, which highlights the importance of storytelling and narratives as key sense-making devices that organize and frame political reality. Examples of such political narratives include stories about power and where it resides, such as online conspiracy theories that position power with an invisible political elite \citep{birchall2022conspiracy}, or stories that shape collective identities, and that pit an ``us'' against an opposing ``them'' \citep{benkler2018network}. As the significance of such narratives for understanding current societal issues becomes increasingly evident, there is a growing demand for methods that support their empirical analysis. Specifically, disciplines across the social sciences and humanities are marked by an urgent need for robust, machine-guided methods that enable large-scale, empirical analysis of political narratives circulating in digital media, including archived political speeches, social media posts, transcripts of parliamentary debates, and political manifestos on party websites. 
This work addresses this need by combining transferable approaches from natural language processing and network science into a theoretically-informed method for extracting relevant \textit{narrative signals} from text.
% We define narrative signals as core aspects of narratives, thereby focusing on actants, their goals, their actions, and their relationships within the narrative. We will refer to the textual instantiations of these signals as ``narrative traces''.
We define narrative signals as core elements of narratives, such as actants, their goals, their actions, and their relationships within the narrative. We will refer to the textual instantiations of these signals as \textit{narrative traces}.

On a methodological level, the study of narratives is typically associated with the practice of \textit{close reading}, whereby selected texts are carefully examined and analyzed by drawing on the conceptual frameworks and sensitizing concepts from the field of (literary) narratology \citep{herman2007routledge,Bal2009}. Recent work in the area of \textit{small stories analysis} has demonstrated that this method of close reading can be fruitfully applied to the analysis of born-digital traces, such as specifically curated collections of messages or posts from social media \citep{georgakopoulou2021small,georgakopoulou2022coopting,page2010reexamining}. Likewise, conceptual advances in transmedial narratology have opened up the field of close narratological analysis to digital media and their emerging storytelling modes \citep{thon2016transmedial}. Yet, as a result of the on-going digitalization of political communication, the volumes of texts that might be considered relevant for studying the narrative aspects of public discourse, have surpassed the capacity of individual researchers to explore each text in depth, and have rendered manual coding and analysis of documents exceedingly expensive \citep{borgman2017big}.

This challenge of information overload highlights the need for alternative, machine-guided reading protocols, which rely on (semi-)automated text processing to surface relevant narrative signals from texts. As illustrated by advances in the interdisciplinary field of \textit{digital humanities} \citep{Burdick2016}, the application of tools and techniques from natural language processing, along with approaches from other computational fields such as (social) network analysis, can inform a practice of \textit{distant reading} \citep{Moretti2013,underwood2019distant} and \textit{guided close reading} \citep{Pournaki2021}, in which fundamental questions in the humanities and social sciences are approached through quantitative analyses of digital corpora. It has thus been shown that sensitizing concepts from narratology, some of which long predating the advent of personal computers, can effectively be operationalized in computational terms \citep{Boyd2020,willaert2023computational}. Likewise, work at the intersections of corpus linguistics and critical discourse analysis has started to explore how corpus-based methods might capture such politically-relevant elements as framing and perspectivization \citep{baker2008useful}. In this paper, we demonstrate that such data-driven approaches to narratives and narrative signals could similarly benefit research in social science and media studies, in particular the study of political discourse on digital platforms. To this end, we propose an unsupervised, computational, network-based approach to extracting and analyzing narrative signals from public discourse.

\subsection{Objectives}

The present paper aims to contribute to the development of robust and replicable methods for the extraction and representation of narrative signals from text to support sociological interpretation of political narratives in large corpora. In accordance with on-going work in digital humanities and computational social science, we approach the extraction of narrative signals from texts as an information-retrieval problem, which we address by combining techniques for natural language processing with methods from network analysis. Specifically, one objective of this paper is to explore how Abstract Meaning Representation (AMR) \citep{Banarescu2013}, a formal language for representing the meanings of sentences, can be leveraged for the study of political narratives. The fundamental methodological challenge our proposed method thus faces, is one of operationalization: namely that due to the complexities of language and discourse, narrative signals, such as references to particular actors and events, might be manifested in texts in many different forms. We argue that this limits the applicability of deductive top-down, pattern-matching approaches that aim to match narrative signals to predefined sequences of words, or to specific syntactic structures. As an alternative to these methods, we therefore propose an inductive, meaning-centric approach, in which we surface relevant narrative signals starting from a representation of a text's \textit{meaning} in the AMR language. More concretely, this paper seeks to make three contributions to the emerging field of corpus-based approaches to political narratives. First, it introduces a graph-based formalism for extracting and representing key narrative signals, based on AMR. These signals are then arranged and re-assembled into networks\footnote{On the use of ``graph'' and ``network'' in this paper: Both words refer to a set of vertices/nodes connected by edges/links. However, in order to avoid confusion between the various outputs of the presented method, we employ ``graph'' when we refer to the objects related to the direct output of AMR (e.g. AMR graph, actor subgraph,...), and ``network'' when referring to objects created through systematic re-assembly of narrative signals (e.g. narrative network, actantial network).} that point to the relevant passages of the text, allowing for the analysis of narratives through ``guided close reading''. Second, the paper offers a computational implementation of aforementioned formalism in the form of open source code which can be reproduced on different corpora. Third, we present an illustrative analysis of a corpus of State of the European Union addresses, by which we demonstrate that the formalism can be used to effectively surface identity-shaping political narratives from public discourse. Before proceeding to discuss our formalism, its implementation, and its applications, we provide an actionable definition of \textit{political narrative} and the specific narrative signals we target in our analysis, and discuss the technical background for our approach.

\subsection{Political narratives: definitions and conceptual foundations}

The recent \textit{narrative turn} in the social sciences \citep{goodson2011narrative} has produced a wide range of definitions and interpretations for the concept of narrative. What connects these manifold interpretations is an acknowledgment that humans rely on narratives as a central meaning-making device through which they make sense of (political) reality \citep{Ranke1967,Fisher1984}. 
Correspondingly, the omnipresent, ``international, transhistorical, transcultural'' nature of narratives has been broadly commented on \citep{Barthes1975}. To reduce some of the complexities associated with this diversity of definitions and interpretations, we opt here for a pragmatic approach, and propose a working definition of political narratives that draws on those previous conceptualizations that are most pertinent to our computational formalism and method.

A first set of relevant definitions can thus be traced back to structuralist approaches to narrative, which, while predating personal computers and the digital tools we currently have at our disposal for processing language and text, have inspired a range of recent machine-guided methods for studying narratives in digital corpora \citep{mani2013computational}. Relaying one of structuralism's core ideas, \citet{Propp1968} proposes that there are only a finite number of archetypical narratives. He thereby defines narratological schemata consisting of specific types of archetypical actors or \textit{dramatis personae}, defined by their function in the plot, according to which all narratives can be analyzed and encoded. Similarly, \citet{Greimas1987} develops \textit{actantial models} in which the entities figuring in narratives are assigned to interrelated actantial roles such has the ``helper'' \textit{versus} the ``opponent'', or of the ``sender'' \textit{versus} the ``receiver''. Comparable actant-centric approaches have been elaborated in the broader field of social science, as exemplified by Bruno Latour's actor-network-theory, which draws on examples from literary studies to include both human and non-human actors (e.g. objects, animals) as part of the analysis \citep{Latour2005}. As such, inductive approaches to actantial modelling have proven productive for the analysis of recent narrative phenomena such as online conspiracy theories, which are marked by complex and rapidly evolving networks of associations between a variety of actors \citep{Tangherlini2020,tuters2022deep,willaert2023computational}. Likewise, we might expect such actor-centric approaches to yield advances in the computational analysis of political narratives in digital media more broadly.

A second body of theoretical concepts from which we draw here, further emphasizes the functions that such political narratives might fulfill. In political contexts, narratives may primarily serve as complexity-reducing devices by explaining potentially multi-faceted phenomena in terms of readily accessible stories \citep{Groth2019}. They typically involve personifications of abstract concepts like ``the economy'' or ``the market'', giving them agency to act upon other concepts. From a cognitive perspective \citep{Bruner1991}, narratives may be approached as sense-making tools that allow individuals and groups to ``understand [themselves] as political beings'' by ``construct[ing] disparate facts in [their] own worlds and weav[ing] them together cognitively'' \citep{Patterson1998}. In this sense, narratives serve as ``organizations of data'' \citep{Fenster2008}: complexity reducing mechanisms that select aspects from the many facets of a given event, and rearrange them into a simpler representation. Through political narratives, actors are thus able to point out their allies and adversaries, and to convey their interpretation of how society should be organized. Narratives, in other words, help political actors form their specific ontologies or worldviews by specifying relations between those elements deemed relevant (e.g. actors, places, events). The selection of specific facets that should be highlighted is one aspect that makes narratives inherently normative \citep{Patterson1998}, as they reveal both the elements that the narrator considers ``noteworthy'', as well as the elements they consider especially salient within an observed political event or process. In this context, narratives furthermore play a pivotal role for the creation of (political) identity. \citet{Somers1994} highlights this identity-shaping feature of what she calls ``ontological narratives'' that presuppose a strong connection between sense of self and agency: it is by positioning themselves in relation to others, that actors might develop a sense of (collective) identity.

Our working definition of narrative is indebted to both these structuralist and cognitive perspectives, and resonates with previous work by \citet{Shenhav2006}, who emphasizes the role of narratives as representations of political reality. The author gives a definition based on the minimal structuralist definition by \citet{Prince1982a}, who states that a narrative is a ``representation of at least two real or fictive events or situations in time sequence, neither of which presupposes or entails the other''. For to Shenhav, \textit{formal} political narratives are narratives that come from political institutions, and \textit{informal} narratives are narratives dealing with political topics. Based on these insights, we propose a slightly modified definition:
\paragraph{Definition}
A \textit{political narrative} is a representation of political reality by means of a sequence of events that are connected temporally and/or causally. Each event involves one or more actors that act upon the world or upon one another. These actors are not necessarily human, they can be personified concepts like ``the market''. Actors are ascribed goals and motives that serve as explanations of their actions within the narrative.

A full-fledged method for extracting political narratives from raw text would then be able to extract the actors involved, their relationships, their ascribed goals, as well as the temporal and causal relationships between the represented events. Applying this to large textual corpora, the method would also be able to discern which elements are part of the core narrative, and which ones are not. In such contexts, however, we observe that political narratives are rarely told explicitly, but that they are much rather \textit{alluded to}. \citet{Polletta2015} argues that these references to narratives are often performed via the introduction of \textit{reference characters}, like the so-called ``welfare queen'', which evokes a specific narrative around welfare abuse. Similarly, certain constellations of actors such as ``growth'', ``jobs'' and ``market'' can be identified as markers for an overarching neoliberal ``meta-narrative'' \citep{Somers1994}. We will refer to such markers as \textit{narrative signals}, and prioritize the extraction of core signals such as actors, events, perspectivizations, and ascribed goals, that allow the researcher to efficiently identify allusions to possible implicit political narratives in textual corpora. In the following section, we discuss previous work from the fields of Natural Language Processing (NLP) and computational text analysis aimed at extracting these specific textual traces (and the relations between them), thus providing an overview of the technical background against which we position our own, network-based and meaning-centric approach to political narratives.

\subsection{Technical background and previous work}

The field of computational narrative understanding has been marked by recent advances in discourse-level approaches to narrative, such as the identification of narrative ``shapes'' including pathways, periods, or arcs, as well as socio-cultural narrative ``schemas'' such as narratives of progress, redemption, or decline \citep{Piper2023}. We base our method for the extraction of narrative signals on a specific line of research that connects narratological theory with network-based computational methods for text analysis. Notably, \citet{Piper2021} argue that data-driven approaches to narrative should surface specific traces of ``narrativity'' from text, which, according to the authors, occurs when ``someone tells someone somewhere that someone did something(s) [to someone] somewhere at some time for some reason'' \citep{Piper2021}. This definition highlights concepts such as agents, events, and perspective, which we likewise place at the center of the present paper. While this means that we momentarily exclude other features associated with narratives, such as causality (aspects to which we will return in our conclusion), we argue that political narratives are rarely explicitly mentioned in political discourse, and that signals such as actor constellations can therefore be sufficient for the researcher to surface the underlying narratives that are being alluded to.

Methods for extracting and representing relationships expressing ``who does what to whom'' from texts, which, along with traces of perspectivization and envisaged goals or motives of actors, form the core object of our analysis, have been explored in a body of previous work that combines techniques from natural language processing with network analysis. Already in his foundational text on ``distant reading'', \citet{Moretti2013} models narratives as networks of interactions between characters (e.g. ``who speaks to whom''). Moretti thereby argues that a key benefit of a network-like representation of narrative is the zoomed out perspective a network visualization affords: ``the idea that the temporal flow of a dramatic plot can be turned into a set of two-dimensional signs - vertices (or nodes) and edges - that can be gauged at a single glance'' \citep[p. 211]{Moretti2013}. Visual inspections of networks thus yield insights into the structure and symmetry of interactions between characters, and centrality measures are used to gauge the significance of specific characters for the structure of the narrative. 
 
One recurring method in the literature on machine-guided narrative analysis, is to extract ``who does what to whom'' relations from text based on syntactical dependency relations, specifically grammatical subject-verb-object relations. Here, an important distinction can be made between approaches relying solely on syntactical relations in text, typically those of subjects, verbs and objects (SVO relations) and approaches based on \textit{semantic role labelling}, which hinge on abstract models of the role a given argument plays in the event described by the predicate \citep{Jurafsky2024}. The former rely on hand-annotations in dependency tree banks, while the latter typically rely on hand-annotations from the Proposition Bank (PropBank) \citep{Palmer2005} on which algorithms are trained to extract semantic roles like ``agent'', ``patient'', ``instrument'' from sentences. Verbs are sense-disambiguated into \textit{frames} that each have their specific sets of possible argument roles. This allows for a distinction between propositions like ``running a bank'' (encoded in the PropBank frame ``run.01'') and ``running to the bank'' (encoded as ``run.02''), which prove useful for the analysis of narratives.

Such extracted relations naturally induce a network between the extracted entities. Various works have exploited these networks in order to approach the underlying narratives present in the corpus. Since the present work follows this line of research, we briefly survey existing network-based methods for the analysis of narratives. While the use of networks as an abstraction is common to most of those approaches, what actually constitutes a narrative differs between them.
\citet{Sudhahar2015} systematically extract SVO relations from a large corpus and statistically determine the most central subjects and objects, as well as their most frequent relations in the text. Labeling these relations as positive \textit{versus} negative, they inductively discover communities of endorsement and opposition in texts about the US elections. \citet{Stuhler2022} uses dependency parsers for sociological inquiry by developing a custom variant of semantic role labeling. This, for instance, allows the author to systematically observe large-scale patterns of interactions between male and female actors, and extract the main semantic motifs associated to the pronoun ``we'' in presidential campaign speeches in the US. \citet{Ash2023} use semantic role labeling to extract the argument structure (``who does what to whom'') from sentences in order to describe relations between entity groups. For the authors, each relation between such grouped concepts (e.g. ``God bless America'') is thereby considered a narrative. \citet{Zhao2024} extract ``narrative fragments'' from a large number of social media posts using SRL and aggregate them into a ``narrative network''. This allows them to identify ``narrative shifts'' in discussions around Covid and the French presidential elections. 

In comparative literature, \citet{Tenen2022} uses semantic role labeling to extract the main characters of a novel and map out their relationships with respect to what the author calls ``relational agency''. Our work is particularly indebted to similar generalist pipelines for the analysis of networks of actors, such as those proposed by \citet{Tangherlini2020} and \citet{shahsavari2020conspiracy}, who have explored their potential for the study of online conspiracy theories. The authors introduce an elaborate scheme combining text embeddings, syntactic parsing and community detection, that allows them to systematically compare the structure of conspiracy theories to real conspiracies. In the field of cultural analytics, work was done by \citet{Bamman2014} in order to scale up the analysis of actantial relations to the level of books \citep{Bamman2014BookNLP}. The authors focus on the extraction of relationships between actors of novels, as well as what these possess and what they are described as being. As we will see, non-human actors can be just as important as human actors in the context of political narratives.

As follows from this concise overview of previous work, the field of network-based narrative extraction from text based on relational approaches is marked by a number of diverging trends. First, on-going research is marked by disparate interpretations of what constitutes a narrative, with some contributions arguing that each connection between two actors might be labelled a ``narrative'', while others assess narratives on the level of larger networks of connections derived from large collections of texts. Similarly, the scope of potential applications of these methods vary, whereby some methods aim to directly implement specific structuralist frameworks, such as actantial models \citep{Finlayson2017,Elfes2024} or narrative arcs \citep{Boyd2020}. Finally, we note that network-based representations for narrative are not restricted to using actors as nodes. Going beyond actor-based networks and focusing on the \textit{event} as the core narrative element, \citet{Bearman2000} model narratives as networks by hand-coding life stories where each node is an event and links designate causal relationships between them.

Our approach aims to complement previous approaches to network-based narrative extraction by abstracting away from syntactic particularities and focusing on the \textit{meaning} of the analyzed sentences. This means that signals of narrative are amplified by flattening the idiosyncracies of different expressions of the same meaning. Furthermore, we provide a replicable pipeline that allows to extract key actors present in the text, their goals and motives, as well as their relationships in the form of networks. These networks can be analyzed at different levels of granularity, either taking into account the full verbs, or collapsing them into broad categories. Through a combination of network analysis and close reading of relevant passages of the text, the underlying political narratives can be reconstructed. This work thereby contributes to the development of robust pipelines for the systematic analysis of narratives in public discourse in a ``guided close reading" framework.
% \vspace{2mm}
\section{Formalism and method}
Drawing on the aforementioned conceptual work from narratology as well as computational work in the fields of text mining and natural language processing, we propose a computational graph-based formalism for identifying and representing selected narrative signals from texts. Rather than to directly operate on the level of strings or syntactical annotations of sentences, our approach starts from a graph-based representation of the meaning of a text according to the Abstract Meaning Representation (AMR) formalism \citep{Banarescu2013}. While the potential of AMR for the study of narratives has already been remarked upon and explored in the context of specific tasks such as text-summarization \citep{Droog-Hayes2018} or biomedical event extraction \citep{rao2017biomedical}, we aim to offer an inductive, robust, and replicable approach that can be broadly applied to surface narrative signals from public discourse. Our main contribution thereby consists of proposing a set of theoretically-informed heuristics for filtering AMR graphs for traces of 1) actors, 2) the events in which these actors are involved, and 3) the perspectivization of events, thus making these narrative traces accessible for further analysis, for instance by representing the relations between actors as networks.

\subsection{Abstract Meaning Representation (AMR)}
\begin{figure}[t]
  \centering
  \includegraphics[width=\textwidth]{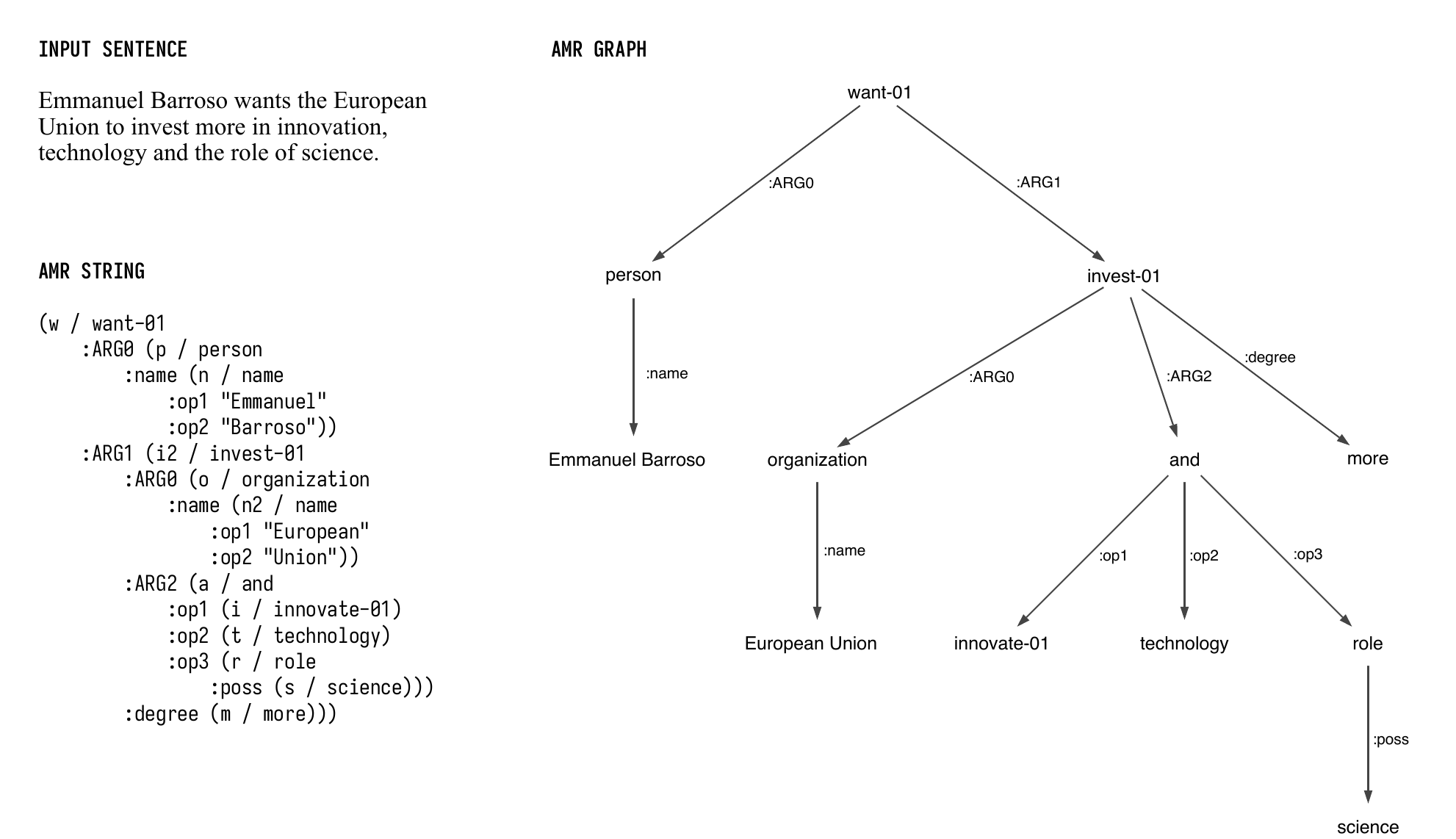}
  \caption{Illustration of AMR representation of the sentence ``Emmanuel Barroso wants the European Union to invest more in innovation, technology and the role of science.'' The AMR-parsed input sentence (left) is represented as a rooted, directed, acyclical graph (right), where nodes are AMR frames and directed edges represent the semantic relationships between them. Name-nodes are labeled by concatenating their op-nodes.}
  \label{fig:amr_example} 
\end{figure}
AMR is a semantic representation language aimed at representing the meaning of entire sentences \citep{Banarescu2013}. As illustrated in Fig.~\ref{fig:amr_example}, this representation language has three defining characteristics. First, AMR's meaning representations abstract away from the specific, syntactic realizations of objects or events. The same meaning representations are thus assigned to sentences that have the same basic meaning, regardless of their formal realization in the text. For example, the sentences ``The physicist made adjustments to the experimental setup.'', ``The physicist adjusted the experimental setup.'' and ``The experimental setup was adjusted by the physicist.'' share the same AMR representation. It is important to highlight here that AMR does not produce word-by-word annotations, but captures over-arching structures of meaning. Second, AMR's meaning representations take the form of rooted, labelled graphs, which can be visualized as such, or written up in PENMAN notation: a serialization format for the directed, rooted graphs used by AMR to encode semantic dependencies \citep{Kasper1989,Goodman2020}. We refer the reader to the Supplementary Material (Sec.~\ref{sec:appendix:penman}) for a description of this format. Third, AMR uses PropBank frames to translate verbs into meaning representations. Those frames thereby capture basic relations between predicates (e.g. ``invest'') and their arguments (e.g. the agent performing the investment, the thing being invested in, or the actor benefiting from the investment).

The example in Fig.~\ref{fig:amr_example} illustrates, on the scale of a single sentence, how AMR might support the task of extracting particular narrative signals from text. First, by abstracting away from actual instantiations in the text, meaning representations offer a flexible approach to the detection and representation of concepts relevant to narrative analysis, notably actors and events. For example, events are not confined to the class of verbs, but might comprise nouns as well (e.g. ``innovate'' and ``innovation'' will refer to the same event). This makes AMR-based approaches less restrictive than rule-based parsers that start from syntactical annotation layers such as dependency tags, and which might require a relatively large set of rules and definitions in order to capture all instances of the desired narrative signal. Second, AMR, by design, represents the meaning of a sentence in terms of a ``who does what to whom'' relation. More precisely, AMR labels the arguments of each predicate in a sentence following the PropBank conventions, thus making the semantic roles of each concept directly accessible for analysis. While these roles are specific to each verb, in general, the ``ARG0'' role thereby corresponds with the ``agent'' role, ``ARG1'' with the ``patient'' role, ``ARG2'' with the instrument role. For previous analyses of textual corpora based on such semantic frames and roles, we refer to \citet{willaert2020building,willaert2022tracking}. Third, AMR graphs can take more complex forms in which argument roles are not only filled by objects, but also by other events. This can be observed in Fig.~\ref{fig:amr_example}: the event of ``investing'' is the patient of ``wanting''. These explicit relations between events open up perspectives for analysing how certain events might be put into perspective by other events, such as events of ``needing'' or ``desiring''. As will follow, this will enable us to surface traces of actors' goals and motives from text. Finally, AMR's graph-based representation resonates directly with the distant reading approaches to the analysis of narratives discussed above, as it allows for a further integration of methods from text analysis with approaches from network analysis. In the following sections, we discuss in more depth how AMR representations can be made suitable for narrative analysis by means of graph-based operations.

\subsection{Graph-based formalization of narrative traces}
\label{sub:graphbasedformalism}

In our graph-based, meaning-centric formalism for the analysis of narrative traces, we represent key aspects of narrative meaning as a series of graphs. To this end, we start by parsing a sentence into an AMR representation from which we then construct a series of graphs that formally represent key narrative signals, specifically references to events, actors, and the relations between events.

\paragraph{AMR graph}
As a preliminary to our formalism, we first construct a full AMR-graph for each sentence in our corpus. This is achieved by first processing each sentence in a text or collection of texts using a state-of-the-art AMR parser \citep{Lee2022}, which yields and AMR string in PENMAN format. We then parse this PENMAN string in order to represent it as a directed acyclical graph (DAG). Following the AMR specification, these graphs represent the meaning of a sentence in terms of relations between AMR ``concepts'', whereby a ``concept'' can be an English word, a PropBank frameset, or a special keyword expressing, among other things, quantities or logical conjunctions \citep{Banarescu2013}. Every node in such a parsed graph is an AMR ``concept'', and links between them denote their relationships in the sentence. Those links can either denote PropBank ARG-relationships, or relationships proper to AMR, like ``:name'' or ``:degree''. Returning to the example in Fig.~\ref{fig:amr_example}, we see that the nodes in the graph do not necessarily correspond to exact words in the sentence. ``Emmanuel Barroso'' for instance turns into two nodes: ``person'' and ``Emmanuel Barroso'', connected by a ``:name''-labeled edge, or ``innovation'' turns into the node ``innovate-01''.

The main contribution of this paper is the translation of AMR-based representations of text into a structured output that allows the researcher to discover narrative signals in the data using straightforward data manipulations. 

The following section describes the different steps involved in this translation task. Formally, given an AMR-parsed sentence, we represent it as a rooted, directed, acyclic graph where vertices are AMR concepts and edges represent their relations. While this graph contains all the information necessary to investigate actors and their relations, some processing steps are useful in order to take advantage of the AMR representation for narrative analysis as well as to facilitate further processing for large corpora. More precisely, the goal is to transform the full AMR graph into a tabular format that can then easily be queried, in order to extract, for instance, the most common actors, or the most common relations between them. We therefore propose a set of formal definitions that guide this subdivision of the full AMR graph into rows that correspond to single events within the sentence. The key idea here is that each event is characterized by a predicate. A sentence may contain several such events. The relationships between the events within the sentence are given by the edges between their predicates. Taking up the example from Fig.~\ref{fig:amr_example} again, there are two events: one given by ``want-01'' and one given by ``invest-01'', while the latter is the patient (ARG1) of the former. The aim is to divide the full AMR graph into rooted subgraphs, where each subgraph's root is a predicate. Formally, we define a \textit{predicate node} as a node with at least one outgoing ARG edge. We now subdivide the full graph into subgraphs, where each subgraph contains only one predicate. We call such subgraphs \textit{event subgraphs}. Then, once the graph is divided, specific sets of vertices are grouped together to \textit{actor subgraphs}. With these processing steps, we are able to transform each sentence into a tabular representation which can then be queried in a traditional manner, and which affords additional analyses of each of the extracted traces.

\begin{figure}[t]
  \centering
  \includegraphics[width=\textwidth]{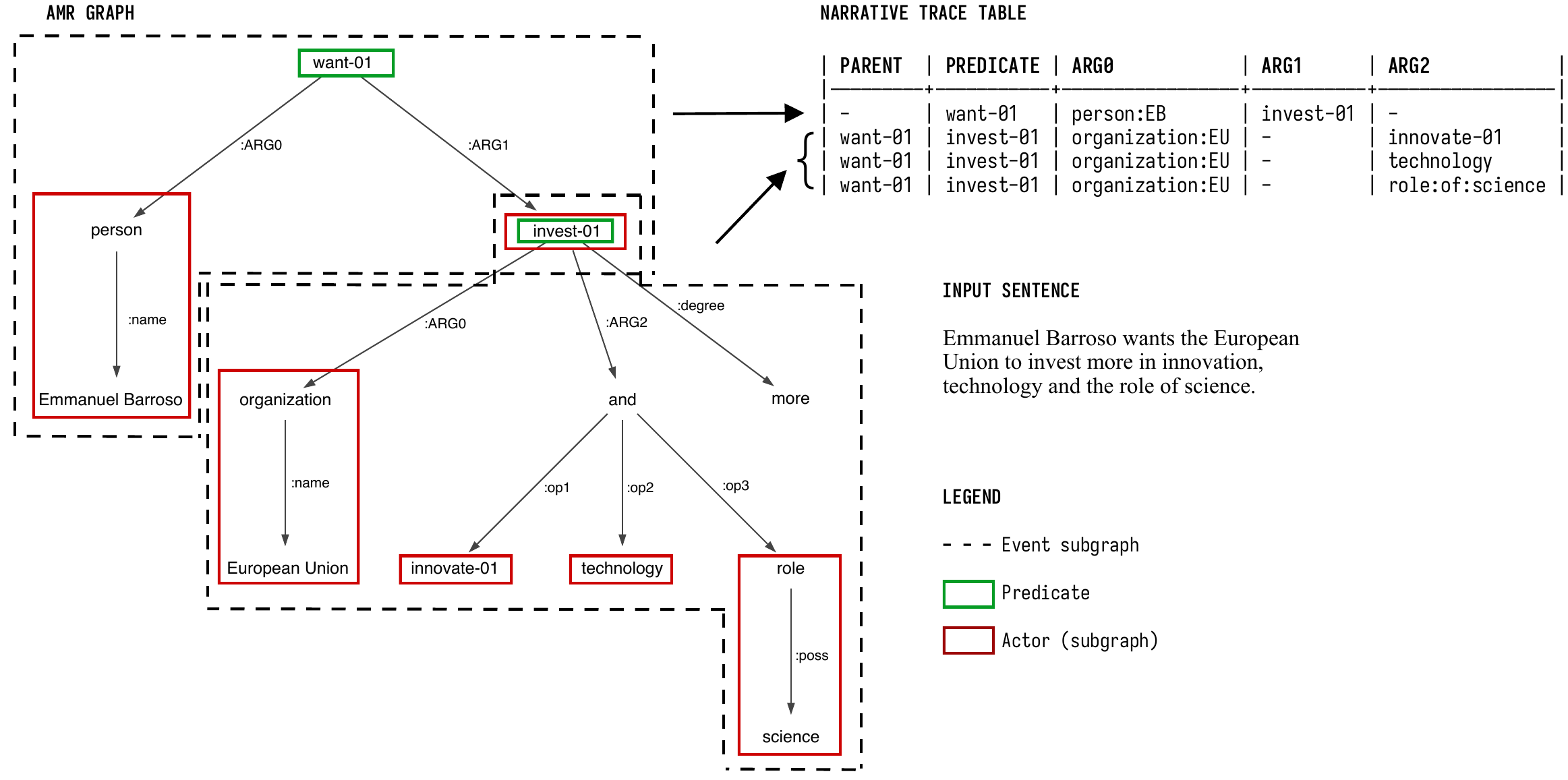}
  \caption{Translation of the AMR graph into structured output from which narrative signals can be extracted.}
  \label{fig:amr_table}
\end{figure}

\paragraph{Event subgraphs}
First, we identify all predicate nodes according to the definition above. For each predicate, we compute all outgoing paths and cut each path as soon as the next predicate is reached. This is when the next event subgraph would start. The current event subgraph is then given by the subgraph induced by the set of nodes present in those paths. In Fig.~\ref{fig:amr_table}, the two event subgraphs are boxed by dotted lines. Note that the node ``invest-01'' is present in both event subgraphs, because it holds the ARG1 position in the first and the predicate position in the second.

\paragraph{Actor subgraphs}
Actor subgraphs are defined for each event subgraph. We define an \textit{actor node} as a node that has an incoming ARG-edge (i.e. is an argument of the predicate) within the event subgraph. We then compute all outgoing paths from a given actor that exist within the current event subgraph. The \textit{actor subgraph} is given by the subgraph induced by the set of nodes present in the kept paths. In Fig.~\ref{fig:amr_table}, the actor subgraphs are denoted by red boxes. Note that the node ``invest-01'' is both a predicate and an actor, depending on which event subgraph is considered.

\paragraph{Narrative trace table}
One central aim of this work is to make AMR representations suitable for the  exploration of narratives in large textual corpora. Therefore, we transform the extracted graphs into a representation that can easily be queried for narrative signals like most common actors, most common actions, goals, or motives. We represent each event subgraph as one tabular line (or more, in the case of ``and'' or ``or'' relations), where column entries correspond to the involved predicates and actor subgraphs. Actor subgraphs are transformed to strings by keeping each involved node's label and concatenating them. 

%% narrative signal = actant, their goals, their actions and relationships in the narrative
%% narrative trace = textual instantiation of a given signal (i.e. a string as we fill it in this table) 

The result for the example sentence can be seen in Fig.~\ref{fig:amr_table}. Note that we show here only a subset of the extracted columns. For an extensive list of columns extracted, we refer the reader to the source code documentation referenced in the software availability section.
As can be seen in Fig.~\ref{fig:amr_table}, AMR encodes ``and''-relations using an ``and''-node that catches the incoming ARG-relation and distributes it to the arguments using outgoing ``op''-edges. In such cases, we add rules to our translator to ensure that the graph is filtered correctly and the ARG-relations are assigned to the correct nodes.

\subsection{Analysis of the narrative trace table for narrative signals}
\label{sec:signalanalysis}

\paragraph{Actors and actions}
Each row in the narrative trace table represents an event, given by a predicate and its arguments. Extracting the most common actors and their actions therefore translates into a value count query on the table. To extract all the events containing actions that a given actor $A$ performs onto actor $B$, the corresponding query in SQL syntax would look as follows:

\begin{verbatim}
SELECT * FROM narrative_trace_table
WHERE ARG0 = "A" AND ARG1 = "B";
\end{verbatim}

Examples of queries reproducing the findings presented Sec.~\ref{sec:results} in are available in the GitHub repository referenced in the software availability section.

\paragraph{Goals and motives}
We use the hierarchical structure that naturally arises from our AMR-based graph formalism in order to extract another narrative signal: the goals and motives of actors. As AMR makes use of PropBank frames, we can readily align them with other ontologies, which allows us to normalize and group predicate frames together into larger categories. In particular, we use the VerbAtlas ontology \citep{difabio2019verbatlas} to group frames into meaningful over-arching categories such as {\sc require\_need\_want\_hope}, which contains frames related to the concept of necessity. Using this higher-level categorization of actions allows us to operationalize the concept of a given actor's goals as follows: first, we identify goal-related verb categories based on the way goals are expressed in the text. With the most common related phrases being ``we need to'', ``we want to'', ``we have to'' or ``we must'', we turn our attention to the following two VerbAtlas categories: {\sc require\_need\_want\_hope} and {\sc oblige\_force}. Then, we generate the list of goals of actor $i$ by taking 1) all events where $i$ holds the ARG0 role and the predicate is a goal verb, and 2) all events where $i$ holds the ARG0 role and the parent predicate is a goal verb. Case 1) allows to extract sentences where the goal the ARG1, such as “He wants ice cream.”, where the goal would be “ice cream”. Case 2) allows to extract more complex propositions such as “He wants to eat ice cream”. The extracted goal here would be “eat ice cream”.

\begin{figure}[t]
  \centering
  \includegraphics[width=\textwidth]{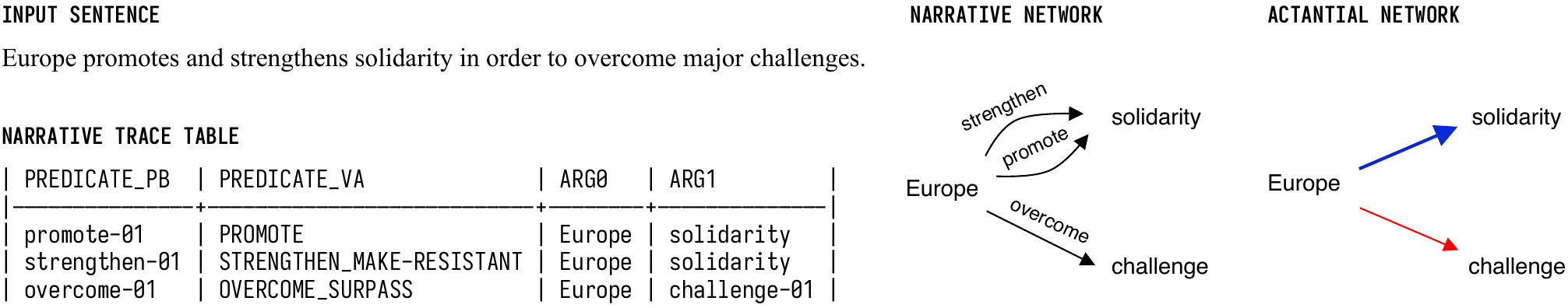}
  \caption{Translation of the narrative trace table into narrative and actantial networks.}
  \label{fig:amr_networks}
\end{figure}

\paragraph{Narrative network}
We transform the narrative trace table into a network, where nodes are actors and links denote the actions they perform onto each other (see Fig.~\ref{fig:amr_networks}). Formally, we can treat the narrative trace table as an edgelist, where the sources correspond to the ARG0s, the targets to the ARG1s and the link labels to the predicates. This generates a directed multigraph $G(V,E)$ that we call \textit{narrative network}. $V$ is an unordered set of nodes and $E$ is a set of ordered links. Each row in the narrative trace table that contains both an ARG0 and an ARG1 generates one link in the narrative network such that the link corresponds to (source, target, label) $=$ (ARG0, ARG1, predicate). For instance, the sentence ``Europe promotes solidarity'' yields the link (``Europe'', ``promote'', ``solidarity''). This network is a simplified representation of the narrative trace table, as it disregards whether two events (given by the predicates) appear within the same sentence, and ignores the hierarchy of the AMR tree. It might also be added that while the use of figurative language such as metaphors is central to political discourse \citep{Mio1997}, our representation of the links adheres to the formal conventions of AMR. Disambiguation of figurative language is thus left to the human interpreter. 

\paragraph{Actantial network}
We further simplify the narrative network by collapsing it into a weighted network, where all links between two given actors are collapsed into a single weighted, scored link. We call this network the \textit{actantial network}.
The aim is to generate a network that encapsules the supportive or conflictive nature of the relationship between two actors. To achieve this, we first label each link in the narrative network as either supportive, conflictive or neutral\footnote{We borrow this nomenclature from Thagard's work on cognitive-affective structures \citep{Thagard2015}}. We approach this by leveraging the VerbAtlas ontology \citep{difabio2019verbatlas}. We manually label the existing VerbAtlas categories according to the following schema: an action or evaluation that agent $i$ performs on patient $j$ is supportive, if it brings benefit to or implies a supportive relationship towards $j$. Examples of such predicates are helping, approving, creating, causing, but also actions that imply a positive relationship between the actors, like hosting or warning. On the other hand, if the action brings detriment to or implies a conflictive relation to actor $j$, we label the corresponding verb category as conflictive. Examples include accusing, attacking, betraying, obliging, preventing or resisting. If the verb does not imply either, then we label it as neutral. Please refer to the Supplementary Material for a complete list of categorized verbs. The link score from node $i$ to $j$ encapsules how supportive, conflictive or neutral the actions of $i$ are towards $j$, taking into account all the links in the narrative network. Let $C_{ij}$ be the matrix that counts the number of links from $i$ to $j$ labeled as conflictive, $S_{ij}$ and $N_{ij}$ the corresponding matrices for supportive and neutral links. We define the link score $\sigma(i,j)$ as
\begin{equation}
\sigma(i,j) = \frac{ S_{ij} - C_{ij}} {S_{ij} + C_{ij} + N_{ij}}
\end{equation}
The link weight between two nodes $i$ and $j$ in the actantial network is given by the sum of links between $i$ and $j$ in the narrative network.

\subsection{Technical implementation}
\label{sec:implementation}

\begin{figure}[t]
  \centering
  \includegraphics[width=\textwidth]{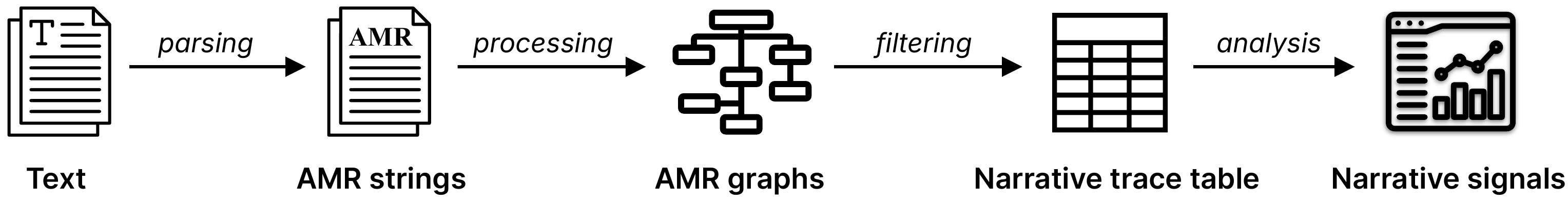}
  \caption{Pipeline for narrative signal extraction using AMR. First, the text is segmented into sentences and parsed to obtain one AMR string per sentence. Then, these strings are processed and transformed into AMR graphs, extracting event and actor subgraphs. These subgraphs are transformed to a tabular representation we call \textit{narrative trace table}, where every row corresponds to one event in the corpus. Finally, the resulting table is analyzed to extract narrative signals.}
  \label{fig:amr_pipeline}
\end{figure}

The full pipeline for the extraction of narrative signals from text is shown in Fig.~\ref{fig:amr_pipeline} and is implemented in Python. We start by feeding the raw text sentence by sentence into an AMR parser. We use IBM's \texttt{transition-amr-parser} \citep{Astudillo2019}, which we run using the pretrained \texttt{AMR2-joint-ontowiki\-seed42} transformer model trained on sentence-AMR annotation pairs \citep{Lee2022}. For each sentence in the corpus, this parser outputs an AMR string in PENMAN format \citep{Kasper1989}. We then use the Python library \texttt{penman} \citep{Goodman2020} to parse the string and transform it to a graph object. All the graph-based algorithms necessary to subdivide the full AMR graph into event, actor subgraphs and transform them into actantial networks are implemented in \texttt{igraph} \citep{Csardi2006}. Finally, we export the narrative trace table as a csv file from which narrative signals can be easily extracted using traditional tabular queries. We provide open source code that performs the analysis from the first to the last step together with this paper. Please refer to the software availability section for links to the respective Github repositories.

\section{Political narratives in State of the European Union addresses (2010 -- 2023)}
\label{sec:results}

We demonstrate how our method can surface meaningful traces of political narratives by applying it to an analysis of a corpus of State of the European Union (SOtEU) addresses. In a process of ``guided close reading'', we explore the extent to which the identified narrative traces can be linked to narratives about European integration previously documented in the literature through qualitative methods. This linkage between traces and contextual knowledge thus relies on human interpretation. The primary objective of this illustrative case study is to offer a demonstration of the presented method. Any in-depth reconstruction of the political agendas of the actors at hand is thus out of the scope of this analysis, nor is it our ambition to ``profile'' politicians based on their utterances. Rather, we are interested in capturing and assessing a diversity of narrative signals, and acknowledge that the speeches of political actors may contain narrative signals that can be associated with various, sometimes conflicting political ideologies. We see this assessment as a precursor to future applications of our method to more elaborate corpora.

The SOtEU is a yearly speech delivered to the European Parliament, in which the President of the European Commission reflects on the past year's achievements, and offers their perspective on the challenges and priorities facing the union in the year ahead. As a corpus of documents that is readily available from the website of the European Union (see the software availability section for data availability), the SOtEU can provide a unique view of the evolution of political discourse, and more specifically on how the European Union as a project has been imagined and re-imagined over time. In this context, narratives play a crucial role in legitimizing European integration \citep{Larat2005,BouzaGarcia2017,Cloet2017,Lueg2020,Blokker2021}. Based on qualitative analyses, \citet{Blokker2021} have identified distinct narratives of European legitimization, each emphasizing the role of different actors and challenges for the European project. We provide a very brief sketch of the identified narratives, some of which we will identify allusions to using the presented AMR-based method. There is the ``rule of law'' narrative, which highlights the importance of legal structures as remedies against authoritarianism; the ``neoliberal narrative'', emphasizing the role of the market as the integrating actor of the EU; the ``federal narrative'' which underlines the importance of non-domination and \textit{shared} sovereignty of the member states; the ``right-wing conservative narrative'' which in turn emphasizes the importance of a common ``cultural identity'' that needs to be safeguarded along with national borders; a ``leftist'' narrative focusing on equality, solidarity and inclusion; finally, a ``participatory'' narrative, in which the EU draws its legitimacy from collective autonomy and an active citizenship engaged in inclusive deliberation. 

Following \citet{Blokker2021}, each of these qualitatively identified narratives are characterized by different archetypal actors. For neoliberal narratives, this includes references to ``the market'', ``growth'' or ``efficiency'', while left-wing socialist narratives are characterized by references to ``equality'',
``solidarity'', or ``inclusion'' \citep[p. 32]{Blokker2021}. \citet{Blokker2021} furthermore identifies the key challenges the EU has to face in each narrative. In the ``participatory democracy'' narrative, for instance, the key challenges identified are depolitization and democratic deficit, the ``right-wing conservative'' narrative's main challenge is the overexpansion of the EU, while the EU must prevent backsliding of the legal system in the ``rule of law'' narrative. The presented method allows us to inductively surface specific constellations of actors from speeches by different presidents of the European Commission, which we then interpret as meaningful signals for some of the briefly sketched political narratives about European integration. Whereas simply counting occurrences of actors skews results in cases such as ``we refuse to follow the logic of the market'', our method makes these relations of support or opposition explicit, thus yielding more fine-grained representations.

Concretely, we examine a corpus of 12 annual SOtEU speeches spanning the period 2010 to 2023. This includes all SOtEU addresses ranging from the first address delivered by Emmanuel Barroso up to and including Ursula von der Leyen's speech in 2023. A generic feature of the State of the European Union speech is that the speaking ``I'' (the narrator) in the speech refers directly to the President delivering the speech.
While it has been demonstrated that pronominal references, particularly the first-person plural, can encompass a wide range of referents \citep{Beard2000,Szabo2022}, as well as entail different levels of inclusion \citep{Pennycook1994,Petersoo2007}, we here assume that ``we'' can broadly be interpreted as referring to the European Union as a whole, given that the speech is delivered in front of the European Parliament.
We can therefore expect the actors ``we'', ``I'', ``European Union'' to play an important part in the underlying ontological narratives. In order to extract meaningful narrative signals alluding to such narratives, we first segment every speech into sentences, and further process the corresponding AMR-representations according to the technical pipeline presented in Sec.~\ref{sec:implementation}. The following sections provide a systematic overview of how the tabular outputs of our method might inform an analysis of different aspects of political narratives present in the SOtEU corpus.
% ----------------------------------
\begin{table}[t]
\caption{Main agents (ARG0), patients (ARG1) and predicates in the SOtEU corpus. Occurrence counts are given in parentheses. Capitalized words correspond to VerbAtlas categories. For the sake of clarity, the words ``this'', ``it'',``that'', ``some'', ``so'' and ``thing'' were removed.}
\label{tab:actors}
\begin{tabularx}{\textwidth}{lX}
\toprule
 & word \\
\midrule
ARG0 & we~(2192), I~(616), Europe~(271), European Commission~(214), they~(183), you~(120), member state~(88), European Union~(83), people~(51), Europeans~(26), country~(25), parliament~(24), other~(21), he~(20), company~(19), she~(18), Greece~(18), house~(16), council~(16), world~(15) \\[1.5cm]
ARG1 & we~(414), Europe~(164), they~(71), economy~(55), I~(48), European Union~(48), climate~(36), job~(33), people~(33), member state~(30), you~(28), more~(26), world~(26), market~(26), society~(25), growth~(25), crisis~(24), finance~(22), euro~(21), treaty~(21) \\[1.1cm]
Predicate & {\sc require\_need\_want\_hope~(695)}, possible-01~(479), {\sc oblige\_force~(425)}, {\sc distinguish\_differ~(415)}, {\sc cause-smt~(356)}, {\sc propose~(303)}, {\sc help\_heal\_care\_cure~(246)}, {\sc affirm~(238)}, {\sc mount\_assemble\_produce~(237)}, {\sc result\_consequence~(201)}, {\sc work~(198)}, {\sc see~(197)}, {\sc give\_gift~(193)}, {\sc exist-with-feature~(189)}, {\sc increase\_enlarge\_multiply~(183)}, {\sc change\_switch~(170)}, {\sc guarantee\_ensure\_promise~(159)}, {\sc ameliorate~(152)}, {\sc protect~(141)}, {\sc agree\_accept~(131)} \\
\bottomrule 
\end{tabularx}
\end{table}
% ----------------------------------
% ----------------------------------
\subsection{Actors and actions}
\begin{table}[t]
\caption{Most representative patient actors (ARG1) for each European Union president. Tf-idf scores are given in parentheses.}
\label{tab:arg1bypresident}
\begin{tabular}{llll}
\toprule
 & Emmanuel Barroso & Jean-Claude Juncker & Ursula von der Leyen \\
\midrule
1 & economy (0.4) & I (0.28) & they (0.39) \\
2 & grow-01 (0.2) & European Union (0.23) & people (0.22) \\
3 & crisis (0.18) & euro (0.14) & Ukraine (0.19) \\
4 & European Union (0.16) & climate (0.14) & climate (0.13) \\
5 & member state (0.16) & job (0.14) & business (0.12) \\
6 & market (0.15) & society (0.13) & I (0.11) \\
7 & they (0.14) & more (0.13) & migrate-01 (0.11) \\
8 & treaty (0.14) & migrate-01 (0.12) & transition-01 (0.1) \\
9 & finance (0.14) & solidarity (0.12) & vaccine (0.1) \\
10 & euro (0.13) & they (0.12) & life (0.1) \\
11 & you (0.12) & European Commission (0.12) & job (0.1) \\
12 & fair-01 (0.1) & time (0.11) & you (0.1) \\
13 & debt (0.1) & Europeans (0.11) & data (0.09) \\
14 & money (0.1) & year (0.11) & world (0.09) \\
15 & banking (0.09) & Greece (0.1) & euro:quant (0.09) \\
16 & competitiveness (0.09) & workers (0.09) & opportunity (0.08) \\
17 & society (0.09) & border (0.09) & energy (0.08) \\
18 & world (0.09) & refugee crisis (0.08) & invest-01 (0.08) \\
19 & bank (0.08) & economy (0.08) & economy (0.08) \\
20 & more (0.08) & Paris (0.08) & NextGenerationEU (0.07) \\
\bottomrule
\end{tabular}
\end{table}
% ---------------------------
% ---------------------------
Table \ref{tab:actors} shows the 20 most frequently identified agents (ARG0), patients (ARG1), and actions (predicates) in the SOtEU corpus. The main actor, both in agent and patient position is the pronoun ``we''. Since the president of the European Commission directs the speech at the European Parliament, this ``we'' refers to a larger actor, to the European Union itself, and it may be employed rhetorically to foster a sense of common identity by grouping together the listeners under a common umbrella. Additionally, the European Union is explicitly mentioned as the third most important actor after the personal pronoun ``I'', which, given the setting in which the SOtEU is delivered, designates the speaker.
The most common patients are, after ``we'' and the European Union, the economy, climate (change), jobs, the market, society. They represent the main themes on which the speeches call the main agents to action. 
The list of most common predicates in the corpus highlights the nature of the SOtEU speeches as projections for the role the European Union will play in the year to come. Predicates related to needs, possibilities, obligations show that the speeches are catered to lay out specific plans and guidelines for the European Parliament. The other verb categories are often related to positive actions, like helping, mounting, affirming, increasing, giving, ameliorating, and guaranteeing. We also observe the mention of challenges and difficulties through the VerbAtlas category {\sc face\_challenge}, which points to another central purpose of the SOtEU speeches: identifying which obstacles are in the way of the speaker's vision for the Union.

We can now segment this broad overview of actors and actions in the full corpus by president in order to compare the narratives each alludes to in their speeches. We choose to focus on actors in the patient (ARG1) position, as they designate the concepts onto which the European Union, or other actors, need to act upon.
In order to systematically compare the prevalence of each patient in each President's speeches, we compute the tf-idf score of each patient concept for all the three Presidents. This score measures the importance of a term in a given subset of the corpus (\textit{document}), which in our case corresponds to the set of speeches of a given president.
If a term appears in all subsets, the score will be lowered, as it becomes less predictive of a given document. High-scoring terms are those with low occurrence across documents and high frequency within a single document. We refer the reader to the Supplementary Material for a definition of the score. The results, computed for the patients extracted from the speeches of each EU President, are shown in Table \ref{tab:arg1bypresident}. We observe a clear shift in emphasis between the presidencies. Barroso highlights the European Union as an economic union, focusing on the financial crisis, the necessity of growth, markets, finance and competitiveness. Comparing this to the narratives identified in \citet{Blokker2021}, we posit that these actors mainly allude to the ``neoliberal narrative'' of the European Union. Jean-Claude Juncker, president between 2014 and 2019, shifts the focus to issues like the job market, migration, calling for solidarity in the European Union, but also highlights the danger of climate change. By highlighting solidarity and the crisis of job markets, Juncker, who is himself aligned with the centre-right European People's Party, provides references to what \citet{Blokker2021} calls the ``leftist narrative'' of European integration. This illustrates that individuals (in particular in a role of President of the Commission), can refer to narratives associated with political ideologies different from their party line. Ursula von der Leyen, finally, sets the focus on the people. Her presidency has seen the Covid pandemic, which is why the vaccine figures prominently in her speech, as well as the Russian invasion of Ukraine. Apart from these events, the main patients show that climate change and migration are on her agenda as well.

While this focus on the president-specific actors shows allusions to narratives that may be explored further by close-reading specific textual passages, the AMR approach allows us now to examine more closely \textit{how} the European Union aims to tackle the aforementioned issues. By extracting the main agents and predicates connected to each patient, we shed light on another narrative signal present in the corpus, namely how the different actors are connected to a larger narrative of European identity. Looking at the most common predicates for ``jobs'' in Jean-Claude Juncker's speeches, we see that the latter mainly speaks about the creation and protection of jobs (``Empowering our economy means investing not just in connectivity, but in job creation.'', SOtEU 2016). Climate change, on the other hand, needs to be fought (``My Commission will work to ensure Europe keeps leading in the fight against climate change.'', 2015), and solidarity shown and expressed (``Secondly, the European Commission, this Parliament and all other Member States will always show loyalty and solidarity with Ireland when it comes to the Irish border.'', 2017). Von der Leyen's speeches explicitly mention the ``rule of law'' narrative, in which people are protected by the law (``The rule of law helps protect people from the rule of the powerful.'', 2020). Furthermore, people might also be pushed into migration (``Honourable Members, Every day, we see that conflict, climate change and instability are pushing people to seek refuge elsewhere.'', 2023.) Ukraine, according to von der Leyen's speech, should be supported (``Our support to Ukraine will endure.'', 2023) and migration managed (``This ultimately comes down to a question of trust... Trust for Europeans that migration can be managed.'', 2021).

\subsection{Goals and motives}
\begin{table}[t]
\caption{Most common predicates associated with goals for the EU. Tf-idf scores are given in parentheses. VerbAtlas categories are in small caps.}
\label{tab:goals:predicates}
\begin{tabular}{llll}
\toprule
 & Emmanuel Barroso & Jean-Claude Juncker & Ursula von der Leyen \\
\midrule
1 & {\sc increase\_enlarge (0.4)} & {\sc show (0.37)} & {\sc move-something (0.32)} \\
2 & {\sc mount\_assemble (0.23)} & {\sc change\_switch (0.27)} & {\sc focus (0.28)} \\
3 & message-01 (0.23) & {do-02 (0.25)} & {\sc continue (0.27)} \\
4 & {\sc go-forward (0.21)} & responsible-01 (0.21) & {\sc work (0.25)} \\
5 & {\sc strengthen (0.2)} & {\sc lend (0.21)} & {\sc guarantee\_ensure\_promise (0.21)} \\
6 & {\sc finish\_conclude (0.2)} & {\sc mount\_assemble (0.17)} & {\sc change\_switch (0.18)} \\
7 & {\sc understand (0.17)} & {\sc remember (0.16)} & {\sc think (0.18)} \\
8 & {\sc combine\_mix\_unite (0.17)} & {\sc know (0.14)} & {\sc see (0.18)} \\
9 & {\sc carry-out-action (0.17)} & union-02 (0.14) & {\sc mount\_assemble\_produce (0.18)} \\
10 & {\sc face\_challenge (0.17)} & {\sc like (0.14)} & {\sc protect (0.18)} \\
\bottomrule
\end{tabular}
\end{table}
% ------------------------------------------------------------------------
Using the approach described in Sec.~\ref{sec:signalanalysis}, we further extract the EU's main goals as evoked by each president. Table~\ref{tab:goals:predicates} shows the most common predicates associated to goals of actor ``we'', ``European Union'', ``European Commission'' and ``Europe'' for each president. Here we see some differences that point to a variety of narratives. Emmanuel Barroso focuses on predicates related to increasing, assembling, producing, finishing and strengthening, while Jean-Claude Juncker focuses on showing, changing and doing. Ursula von der Leyen's main used predicates in relation with goals are those related to moving, focusing, continuing, working and promising. Looking more closely at some of the goals related to those predicates, we find for Barroso goals are related to growth, catering to the neoliberal narrative, ``deepen[ing] economic coordination and integration'' (SOtEU 2011) and ``intensify[ing] our engagement with international partners'' (2010). Barroso further puts forward goals to ``complete the internal market of energy'' (2010), ``complete the monetary union with real economic union'' (2011), ``complete the single market'' (2012) and ``complete connecting Europe'' (2013). Jean-Claude Juncker focuses on the actions the EU needs to show, for example ``show[ing] Russia the cost of confrontation'' (2015), ``show[ing] that we can be fast and decisive on the things that really matter'' (2016) or ``show[ing] solidarity with Africa'' (2017). Juncker furthermore highlights the necessity of taking responsibility ``for protecting our interests and the European way of life'' (2016), ``in recognising when some decisions are not for [the European Commission] to take'' (2016) when member states cannot come to agreements, which is an instance of the ``federal narrative'' identified in, by which the EU sovereignty is shared across member states.
Von der Leyen wants the EU to ``rebuild the trust amongst us and move forward together'' (2020), to focus ``in our investment on professional education and upskilling'' (2022), ``on cooperation with legitimate governments and regional organisations'' (2023), ``on ... technology - and in particular artificial intelligence'' (2020), ``investments on secure connectivity, on the expansion of 5G, 6G and fiber'' (2020). Furthermore, the EU should ``secure [industrial] data for Europe and make it widely accessible'' (2020), ``make sure that European citizens and those around the world have access to [the Covid vaccine]'' (2020), ``ensure there is no [rule of law] backsliding [in EU member states]'' (2020), or ``ensure [the EU's] security of supply and, at the same time, ensure [the EU's] global competitiveness.'' (2022). In her expressed goals, we therefore observe allusions to both the rule of law and the neoliberal narrative.

Turning again to the most common predicates in Table~\ref{tab:goals:predicates}, we see that {\sc mount\_assemble\_produce} is prevalent across EU presidents. We therefore show the different goals related to this action in Table~\ref{tab:goals:mount}. We see that the projects for the European Union range from building energy grids, establishing a unified European labour market, a banking and fiscal union to fostering environments that attract investments, building partnerships with Africa, to setting up a European Health union and rebuilding trust. This example shows how the AMR-based method can be used to easily extract specific goals of given actors and connect them to potential implicit narratives.
% ====================================================================================
\begin{table}[t]
\caption{European goals within the verb category {\sc mount\_assemble\_produce}.}
\label{tab:goals:mount}
\begin{tabular}{lllr}
\toprule  
Predicate & ARG1 & Speaker & Year \\
\midrule
build-01 & energy grids & Barroso & 2010 \\
create-01 & European labour market & Barroso & 2012 \\
create-01 & fiscal and banking union & Barroso & 2012 \\
create-01 & framework that attracts investors & Juncker & 2016 \\
create-01 & environment to invest in & Juncker & 2016 \\
build-01 & partnership with Africa & Juncker & 2018 \\
create-01 & framework that brings investments to Africa & Juncker & 2018 \\
make-01 & Europe's Digital Decade & von der Leyen & 2020 \\
create-01 & vitality & von der Leyen & 2020 \\
rebuild-01 & trust & von der Leyen & 2020 \\
build-01 & European Health Union & von der Leyen & 2020 \\
build-01 & foundation for collective decision-making & von der Leyen & 2021 \\
\bottomrule
\end{tabular}
\end{table}

% ------

\subsection{Actantial networks}
\begin{figure}[t]
  \centering
  \includegraphics[width=\textwidth]{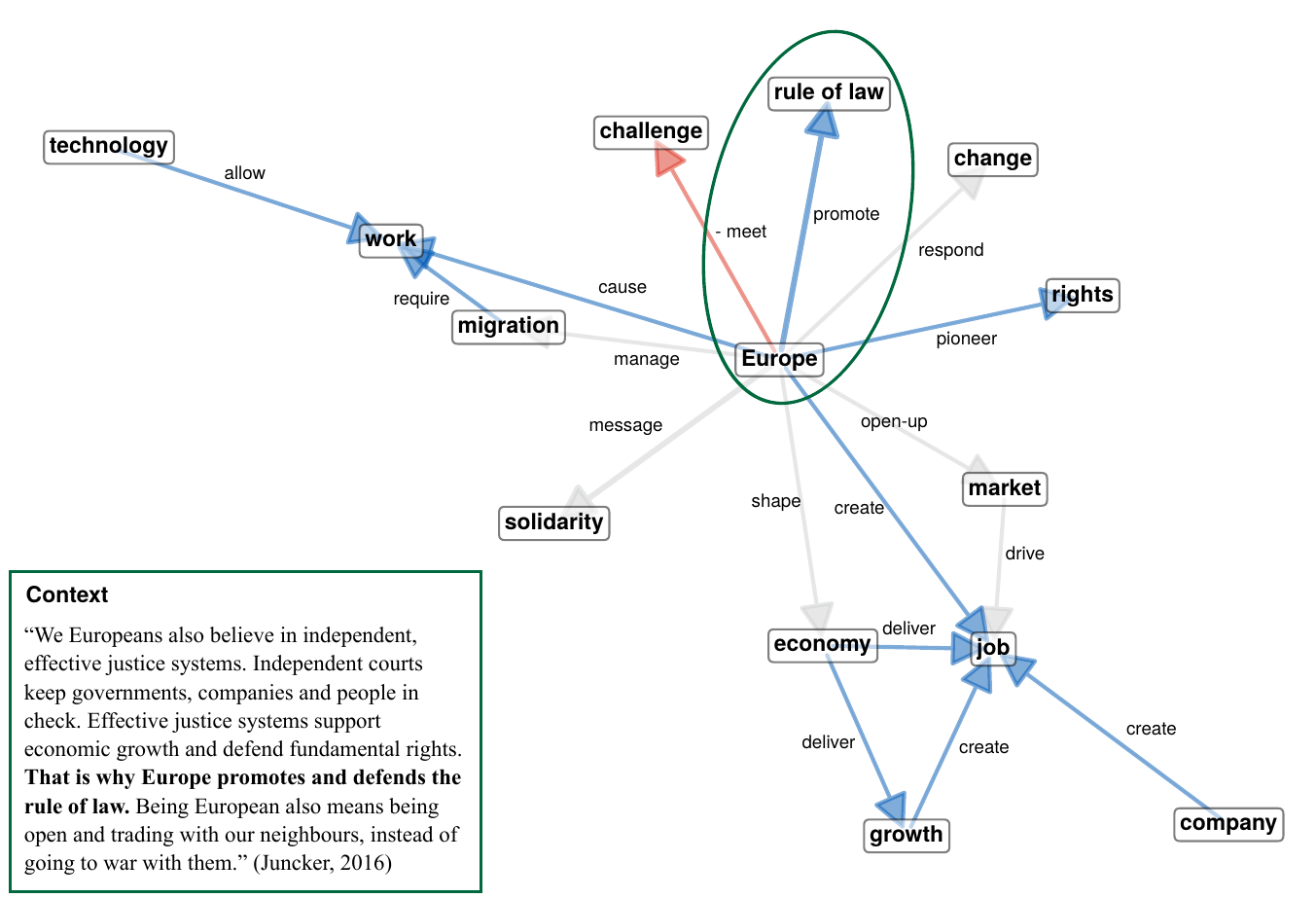}
  \caption{Subset of actantial network of SOtEU speeches. Nodes were selected based on their degree and betweenness centrality. Blue/red/grey links correspond to supportive/conflictive/neutral relationships. The inset demonstrates the guided close reading approach for the selected link from ``Europe'' to ``rule of law''.}
  \label{fig:actantialnetwork}
\end{figure}
\begin{figure}[t]
  \centering
  \includegraphics[width=\textwidth]{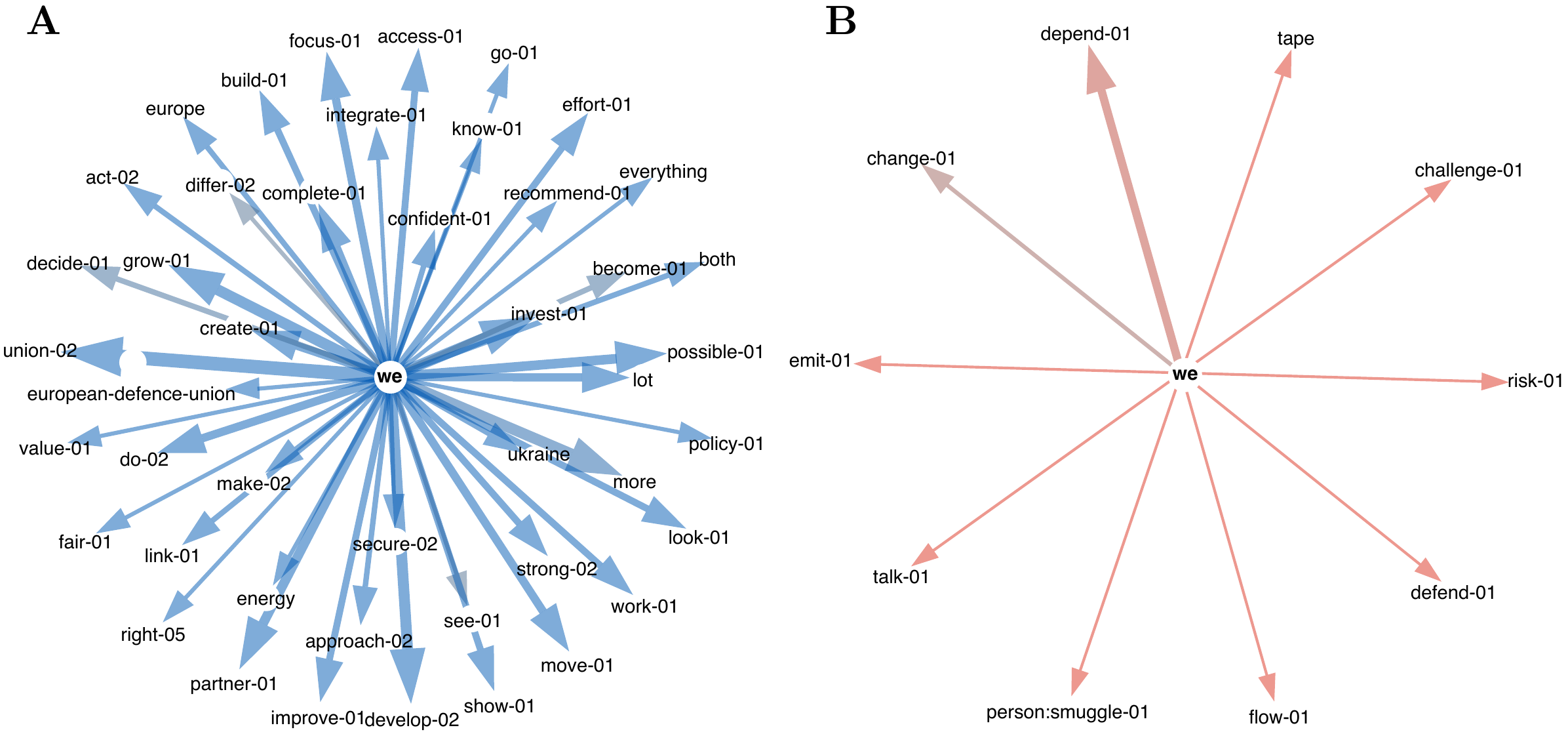}
  \caption{Ego networks of the actor ``we'', displaying positive (\textbf{A}) and negative edges (\textbf{B}). For reasons of graph legibility, only edges with $w(i,j) \geq 3$ are shown in \textbf{A} and $w(i,j) \geq 2$ in \textbf{B}. Edge width denotes weight.}
  \label{fig:egographs}
\end{figure}
In the previous two sections, we have demonstrated the ability of the AMR-based method to transform a textual corpus into a queryable table from which the main actors, their actions, as well as their goals can be extracted using simple tabular operations. We now shift to another data representation in order to shed light on a typical feature of political narratives, namely the identification of protagonists and antagonists. Our approach allows to extract these by examining the types of actions a given actor exerts on others, and by representing these relations as a network, which we refer to as an \textit{actantial network}. For the speeches at hand, the actantial network is made of 1518 nodes and 1781 links. We identify the central actants in the narrative by sorting the list of nodes in the actantial network by degree centrality (sum of in- and out-degree) and betweenness centrality. Looking at these central nodes we find some the main actants presented in Table~\ref{tab:actors}, such as  ``we'', ``I'', ``Europe'', ``economy'', ``growth'', ``job''. Figure~\ref{fig:actantialnetwork} shows a selected subset of the actantial network. Link color reflects the score of the link: supportive, conflictive and neutral relationships are denoted by blue, red and grey colors respectively. We observe traces of the neoliberal narrative with the supportive triangle of ``economy'', ``job'' and ``growth'' (``Growth is the lifeblood of our European social market model: it creates jobs and supports our standard of living.'', 2012). Furthermore, we observe a reflection of Europe's promotion of the rule of law, as highlighted in the inset of Fig.~\ref{fig:actantialnetwork}.

To shed light on the narrative traces related to the collective European identity, we focus on the ego network of the actant ``we'' depicted in Fig.~\ref{fig:egographs}. 
Examples of positively tied concepts are values, access, development, energy, integration, Ukraine, investment, security, or the European Defence Union. On the negative side, we see nodes like challenges, smugglers, dependence, risk, radicalization, fragmentation, and emission. Note that some links in the presented ego network are not immediately intelligible, like the negative link between ``we'' and ``tape''. Examining the sentences that led to this link reveal that the EU aims to reduce \textit{red tape}, which refers to bureaucratic procedures. This example points to some limitations of the presented approach that will be further discussed in the following section. For one, it shows that the the used AMR parser can only parse concepts that it encountered during the training process, and we can assume that ``red tape'' is not part of it. Furthermore, it shows that such missing elements could be resolved in certain cases by including adjectives into the actor representation. While AMR does not formally encode parts of speech that would allow us to identify adjectives directly, PropBank framesets contain this information. The method could therefore be extended to identify adjectives and add them to the actor representations.

Finally, we emphasize that the presented actantial network is a very first step towards a more extensive network-based representation of the narrative signals extracted from the narrative trace table. We envisage such (interactive) network representations as a text-corpus exploring device, connecting distant and close reading in a similar fashion as existing \textit{observatories} \citep{Pournaki2021}, allowing the researcher to seamlessly access the actual textual passage that left the observed narrative trace. A first approach at such an interactive observatory, displaying the full actantial network extracted from SOtEU speeches, can be found here: \url{https://pournaki.com/demos/soteu-narratives/}.
% \url{https://pournaki.com/demos/soteu-narratives}.

\section{Findings and discussion}

Different political actors might use public discourse to propagate varied narratives about political reality, highlighting distinct networks of relevant actors and the events in which they partake, as well as specifying their future goals and objectives. Neoliberal narratives for instance focus on markets and economic growth, whereas more left-wing narratives emphasize solidarity and inclusion. In our case study of State of the European Union addresses (2010 -- 2023), we have demonstrated that key signals for such political narratives, including references to actors, events, and the relations among them, can be extracted from digital corpora by means of an AMR-based formalism and method.

The meaning-centric, network-based formalism introduced here offers a number of advantages over established methods for the analysis of narrative traces. Specifically, when we compare our approach to related methods for extracting ``who does what to whom'' relations from text, our formalism provides direct access to a fine-grained meaning representation of actors and events: in addition to core frames and semantic roles, the AMR-graphs on which our formalism is based allow for the examination of negation, entity types, modality, and other relevant linguistic information, such as the specific names that might be used to refer to an entity. Detailed, meaningful traces of events and actors, as well as the relations among these traces (such as events that are arguments of other events), can furthermore be extracted from AMR-graphs by means of a small number of relatively simple heuristics. This is a direct result of how AMR abstracts away from the grammatical manifestation of a sentence and normalizes meaning, offering for example an identical representation of the same event regardless of whether it is referred to as a noun or a verb in the original text (``innovation'' is, for instance, parsed into the PropBank frame ``innovate-01''). AMR's use of PropBank frames furthermore affords a high degree of interoperability with other ontologies and knowledge bases, such as the VerbAtlas ontology used here to classify predicates. Such classifications and ontologies allow for further normalization of meaning representations and further reduce the number of heuristics or queries that need to be developed in order to extract and study traces of narratives. Finally, a key advantage of starting from AMR is that the formalism by definition represents meaning in a graph-based format, which befits the network-based approaches that are central to the emerging field of \textit{distant reading} and the computational analysis of narratives.

One of the key contributions of this paper is that it introduces an actionable formalism and a computational method for making AMR work for the study of narratives. As such, this paper, to our knowledge, is amongst the first to systematically bridge gaps between two established, yet thus far mainly distinct research traditions. For one thing, we have drawn on insights from narratology and previous conceptualizations of political narratives in order to highlight actors, events, and the relations between actors and events as tell-tale traces of underlying narratives. Central concepts such as actantial networks, which inform the analyses outlined in our case study, can thereby be traced back to structuralist approaches to narratology. While such concepts predate corpus-driven approaches to narrative, we inscribe our work within a line of research that is marked by the rediscovery and further computational operationalization of these foundational ideas. For another, on the side of language processing, our method starts from an approach to language that centers on sentence semantics. Abstract Meaning Representation comes from a tradition of semantic parsers trained on so-called sembanks: pairs of sentences and their corresponding semantic representations. Specifically, AMR-based semantic parsers such as the one used for this paper are trained on datasets of sentences with corresponding representations of their logical meanings. Processing texts with state-of-the-art models for AMR-parsing thus provide the researcher with direct access to a formal representation of the meaning of each sentence in a corpus. %\citep{tosik2015abstract}.

By aligning narratological concepts with an AMR-based formalism, this paper has contributed a robust and replicable method for extracting selected narrative traces from public discourse. However, as will be reflected upon in our conclusion, more work remains to be done to both expand the range of narrative signals to be extracted, as well as to facilitate the further interpretation of such signals, in particular when exploring larger corpora.

\section{Conclusions and avenues for future research}

In this paper we have proposed a network-based approach to the extraction of narrative signals from public discourse, based on Abstract Meaning Representation (AMR). The working definition of political narratives from which we set out, is that political narratives are representations of political reality in the form of a sequence of causally or temporally connected events. Each of these events thereby involves one or more (human or non-human) actors which, driven by specific motives or envisaged goals, act upon the world or upon one another, thus inducing state changes. 

While our formalism and method allow us to extract core narrative signals from text, it is evident that some aspects our initial definition remain outside of the scope of the present approach. First, we only extract explicit relations between events in cases where one event is an argument of another within one sentence. Our method is therefore not yet suited for capturing the implicit temporal or logical relations between events, such as suggestions of causality. Next, the method as such does not keep track of how the states of certain actors might change in the course of a narrative. Finally, the method in its current form only captures explicit references to actors' goals and aspirations, while an actual attribution of underlying drives and motives to actors might again be dependent on prior knowledge of the world. To this, it should be added that this paper has only started to explore how the extracted narrative signals might be interpreted and contextualized further, for now limiting such analyses to querying the narrative signal table as well as a visual inspection of the actantial network. Therefore, by way of conclusion, we identify a number of avenues for future research through which our proposed method and formalism might be extended and improved. We thereby highlight possible technical elaborations of the method itself, and envisage three ways in which our approach might be incorporated into more elaborate machine-guided reading protocols.

On a technical level, our method currently starts from sentence-by-sentence AMR representations of a text. As a result, intra-textual links such as coreferentiality are currently left out of the scope of the analysis, leaving pronouns such as ``it'' or ``they'' unresolved. The outputs of our method might thus be improved upon by first preprocessing the text using techniques and parsers for coreference resolution, and then parsing the resolved texts using the formalism discussed here. This might yield more precise meaning representations, and correspondingly more informative narrative signals. Likewise, our method might benefit from further exploring the potential of the different semantic relations specified by the AMR standard. A further refinement of the formalism might thus also take into account those additional semantic relations, such as quantifiers, expressions of degree, or subsets, allowing for more fine-grained analysis of texts on the level of narrative meaning. Another important limitation in this context is that the AMR formalism is mainly geared towards English \citep{Banarescu2013}. While there exist approaches to AMR for other languages, like Chinese, German, Italian and Spanish \citep{Lee2022}, the majority of implementations for parsing focus on English.

Turning our attention to the network-based representations introduced in this work, we note that the actantial network drastically reduces the complexity of the narrative network (see Fig.~\ref{fig:amr_networks}) by collapsing all relations from one actor to another into a single score capturing the supportive or conflictive nature of the relationship. While is a useful first approach to reduce the complexity of the narrative network, we may consider more fine-grained distinctions that would for example allow to distinguish evaluations from actions. Furthermore, the presented approach to label links in the actantial network depends solely on a context-free categorization of verbs. While this yields meaningful results in clear, non-idiosyncratic speech, it may fail in more complicated cases, where the nature of the implied relationship is context-dependent. Consider for instance the sentence ``The surgeon cut the patient.''. Cutting someone implies a conflictive relationship, but not within the presented medical context. Similarly, there can be context essential to the inference of the nature of the implied relationship spilling into the next sentence. Consider the sentence ``I got the vaccine, even though I am scared of needles. You should get it too!''. The relationship of \textit{getting} on its own is neutral, but in the context of the following sentence it implies a supportive relationship from \textit{I} to \textit{vaccine}. Such context-dependent cases can be approached with the help of LLMs, as has been shown as a first proof of concept in \citet{Pournaki2025a}. To conclude on the potential improvements of the actantial network, we highlight the necessity of generalizable methods for merging nodes that refer to the same actant (e.g. instances of "EU" and "European Union"). Approaches to solve this problem exist for protagonists in literary texts using regular expressions \citep{Bamman2014} and general named entities using external knowledge bases \citep{Cucerzan2007}. Extending and incorporating such approaches into the present pipeline for the analysis of political narratives across corpora is subject of future work. Going further, an abstract categorization of actors similar to the verb categorization presented in this work would enable the analysis of the narrative networks at different levels of granularity, for example by collapsing all nodes that refer to politicians, or laws. This could be a first step to analyze different levels of (meta-)narratives. 

Further methodological gains can be expected from integrating our method into more full-fledged ``observatories'' and machine-guided reading protocols that might facilitate the interpretation of narrative signals \citep{willaert2020building,Pournaki2021}. In particular, the tabular output format currently presented here could be supplemented with more advanced network visualization and exploration tools, allowing for more interactive explorations of a given dataset. Sensible aggregation methods need to be developed for such networks in order to make them maximally useful for narrative analysis. Likewise, we can expect that the scope of narrative traces that can be extracted through this method might be expanded by establishing connections with external knowledge bases \citep{blin2023okg}. This might especially benefit the extraction of narrative signals such as causal relations between events, which hinge on implicit knowledge about the world. In addition to structured ontologies and other knowledge graphs, this future line of research might benefit from exploring the possibilities of large language models (LLMs), where context windows are not limited by sentence boundaries. While the use of LLMs could for instance allow for the identification of implicit causal relationships between narrative events across sentences, the scalability and validation of such approaches raises methodological concerns that need to be carefully assessed. Likewise, further methodological gains can be expected from connecting the empirical traces that are uncovered through the present approach with computational models of how narratives emerge at the intersections of empirical observations and previous knowledge \citep{Vaneecke2023}.

Another path along which the method might be further improved and refined, is through more elaborate empirical case studies that might comprise more idiosyncratic narratives and narrative signals. One active area of investigation here concerns the study of online disinformation and conspiracy theories that have been shown to challenge common-sense conceptions of what might constitute a narrative. Such objects of study in particular call for inductive approaches to narrative, which would have to be developed further in order to deal with the particular modes of storytelling that characterize social media. In this context, the presented formalism could be suited to investigate what has been known in the literature as the \textit{epistemological crisis}, after which antagonistic groups ``may hold different set[s] of interpretive frameworks that determines how and what [they] see of reality'' \citep{Friedman2023}. We argue that narratives are an integral part of such interpretive frameworks, and their analysis and comparison on large-scale corpora of digital trace data may yield important insights that help better understand democracy-endangering phenomena like polarization and issue alignment \citep{Pournaki2025a}.

\section*{Data and software availability}
\label{sec:software}
The SOtEU addresses analyzed in this paper were downloaded from \url{https://state-of-the-union.ec.europa.eu/index_en} and made available as structured machine-readable data under \url{https://github.com/pournaki/soteu-dataset}. For parsing the speeches, our optimized fork of IBM's \texttt{transition\-amr-parser} is available here: \url{https://github.com/pournaki/transition-amr-parser}. The code that transforms the raw AMR strings into tables allowing to reproduce the results presented in this paper is available under \githubrepo.

\section*{Acknowledgements}
AP and TW wish to thank Eckehard Olbrich, Jean-Philippe Cointet, and Thierry Poibeau for their valuable comments and suggestions, as well as Jürgen Jost for fruitful discussions.
\vspace{-3mm}
\paragraph{Funding}
AP and TW acknowledge funding by the European Union’s Horizon Europe programme under grant agreement ID 101094752: Social Media for Democracy (SoMe4Dem) - Understanding the Causal Mechanisms of Digital Citizenship. AP acknowledges funding by the French government under management of Agence Nationale de la Recherche as part of the ``Investissements d'avenir'' program, reference \mbox{ANR-19-P3IA-0001} (PRAIRIE 3IA Institute). TW acknowledges the Flemish Research Council (FWO) for generously funding a long research stay at SciencesPo médialab (Paris) during which the foundations for the present work were established. The funders had no role in study design, data collection and analysis, decision to publish, or preparation of the manuscript. 

\section*{Ethics and personal data}
No user information was processed for the present research, the SOtEU speeches analyzed in the paper are in the public domain.

\begingroup
    \renewcommand\bibname{References}
    % \setstretch{1}
    % \setlength{\bibsep}{5pt}
    % \bibliography{bibliography.bib}
    \printbibliography
\endgroup
\newpage
\appendix
\section*{Appendix}
\renewcommand{\thesubsection}{A.\arabic{subsection}}
\renewcommand{\thefigure}{A\arabic{figure}}
\setcounter{figure}{0}
\renewcommand{\thetable}{A\arabic{table}}
\setcounter{table}{0}

\subsection{tf-idf computation}
\label{sec:appendix:tfidf}
The tf-idf score is a heuristic score that measures the discriminative nature of a word for a given document. Let $C$ a $m \times n$ matrix where $m$ is the number of words and $n$ is the number of documents. The entry $C_{ij}$ gives the number of times word $i$ appears in document $j$. Similarly, we define the matrix $O$ where the entry $O_{ij}$ is 1 if the word $i$ occurs in document $j$ and 0 otherwise.
For each word $i$ and document $j$, we compute the so-called tf-idf score $\mathrm{tfidf}(i,j)$ given by

\begin{equation}  
  \mathrm{tfidf}(i,j) = \mathrm{tf}(i,j) \cdot \mathrm{idf}(i,j)
\end{equation}

where

\begin{equation}
  \mathrm{tf}(i,j) = \frac{C_{ij}}{\sum_i C_{ij}}
\end{equation}

and 

\begin{equation}
  \mathrm{idf}(i,j) = \log \frac{n}{\sum_j O_{ij}}
\end{equation}

For the computation of tf-idf scores in Table~\ref{tab:arg1bypresident}, a word is an AMR concept holding the ARG1 role and a document is given by grouping all speeches of one president. For Table~\ref{tab:goals:predicates}, a word is a predicate associated to a goal and a document is again given by grouping all speeches of one president.

\subsection{PENMAN notation}
\label{sec:appendix:penman}
% \begin{verbatim}
% ;    ┌────────────────────────── Variable (this one is the graph's top)
% ;    │     ┌──────────────────── Instance relation
% ;    ┴ ────┴─────
%     (d / drive-01
% ;      ┬ ───┬────
% ;      |    └─────────────────── Concept (node label)
% ;      └──────────────────────── Indicates the node's concept
% ;            ┌────────────────── Edge relation
% ;      ──────┴───────
%        :ARG0 (h / he)
% ;      ──┬──
% ;        └────────────────────── Role (edge label)
%        :manner (c / care-04
% ;                      ┌──────── Attribute relation
% ;                 ─────┴─────
%                   :polarity -))
% ;                           ┬
% ;                           └─── Atom (or "constant")
% \end{verbatim}

PENMAN notation was originally concieved by \citet{Kasper1989} as a serialization format for the directed, rooted graphs used to encode semantic dependencies. The following diagram taken from \citet{Goodman2020} breaks down the elements of the PENMAN notation for the AMR representation of the sentence ``He drives carelessly.'':

\begin{figure}[h]
\includegraphics[width=.6\textwidth]{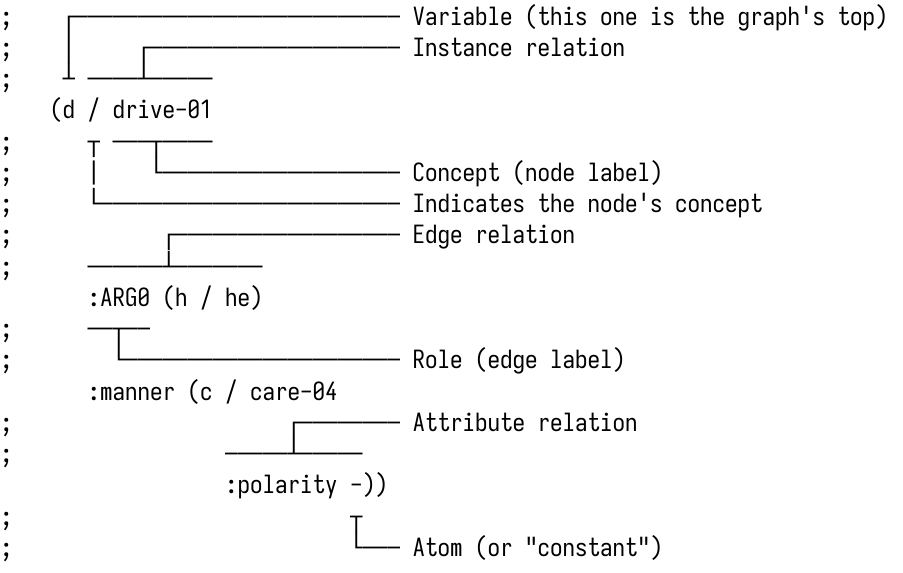}
% \caption[]{\label{fig:penman} Penman example.}
\end{figure}

\subsection{Verb annotations}
\label{sec:appendix:verbs}
Each VerbAtlas frame was annotated by the two authors individually as either beneficial, adverse or neutral. The results were then merged and the few inconsistencies resolved in mutual discussions. We report the results for beneficial and adversely labeled verbs here:
\paragraph{Supportive VerbAtlas frames}
{\sc accompany}, {\sc achieve}, {\sc adjust\_correct}, {\sc affirm}, {\sc agree\_accept}, {\sc ally\_associate\_marry}, {\sc ameliorate}, {\sc apply}, {\sc approve\_praise}, {\sc argue-in-defense}, {\sc ascribe}, {\sc authorize\_admit}, {\sc befriend}, {\sc believe}, {\sc benefit\_exploit}, {\sc carry\_transport}, {\sc cause-smt}, {\sc celebrate\_party}, {\sc co-opt}, {\sc come-from}, {\sc commune}, {\sc compensate}, {\sc court}, {\sc create\_materialize}, {\sc earn}, {\sc embellish}, {\sc enjoy}, {\sc fit}, {\sc follow\_support\_sponsor\_fund}, {\sc generate}, {\sc give-birth}, {\sc give\_gift}, {\sc group}, {\sc grow\_plow}, {\sc guarantee\_ensure\_promise}, {\sc harmonize}, {\sc have-sex}, {\sc help\_heal\_care\_cure}, {\sc hire}, {\sc host\_meal\_invite}, {\sc incite\_induce}, {\sc increase\_enlarge\_multiply}, {\sc lead\_govern}, {\sc liberate\_allow\_afford}, {\sc lighten}, {\sc like}, {\sc mount\_assemble\_produce}, {\sc nourish\_feed}, {\sc obey}, {\sc offer}, {\sc pardon}, {\sc pay}, {\sc preserve}, {\sc promote}, {\sc protect}, {\sc prove}, {\sc raise}, {\sc rely}, {\sc renew}, {\sc repair\_remedy}, {\sc require\_need\_want\_hope}, {\sc result\_consequence}, {\sc satisfy\_fulfill}, {\sc settle\_conciliate}, {\sc share}, {\sc simplify}, {\sc stabilize\_support-physically}, {\sc start-functioning}, {\sc strengthen\_make-resistant}, {\sc take-shelter}, {\sc tolerate}, {\sc warn}, {\sc wash\_clean}, {\sc welcome}, {\sc win}

\paragraph{Conflictive VerbAtlas frames}
{\sc abstain\_avoid\_refrain}, {\sc accuse}, {\sc attack\_bomb}, {\sc betray}, {\sc bewitch}, {\sc blind}, {\sc break\_deteriorate}, {\sc burden\_bear}, {\sc cage\_imprison}, {\sc cancel\_eliminate}, {\sc castrate}, {\sc compete}, {\sc conquer}, {\sc contract-an-illness\_infect}, {\sc corrode\_wear-away\_scratch}, {\sc corrupt}, {\sc criticize}, {\sc cry}, {\sc debase\_adulterate}, {\sc deceive}, {\sc defeat}, {\sc delay}, {\sc destroy}, {\sc dim}, {\sc dirty}, {\sc disappear}, {\sc disband\_break-up}, {\sc discard}, {\sc dislike}, {\sc dismiss\_fire-smn}, {\sc downplay\_humiliate}, {\sc drive-back}, {\sc endanger}, {\sc exhaust}, {\sc face\_challenge}, {\sc fail\_lose}, {\sc fake}, {\sc fight}, {\sc flatter}, {\sc forget}, {\sc frustrate\_disappoint}, {\sc give-up\_abolish\_abandon}, {\sc hit}, {\sc hunt}, {\sc hurt\_harm\_ache}, {\sc isolate}, {\sc kill}, {\sc knock-down}, {\sc leave-behind}, {\sc lose}, {\sc lure\_entice}, {\sc mess}, {\sc miss\_omit\_lack}, {\sc mistake}, {\sc oblige\_force}, {\sc offend\_disesteem}, {\sc oppose\_rebel\_dissent}, {\sc overcome\_surpass}, {\sc preclude\_forbid\_expel}, {\sc pretend}, {\sc punish}, {\sc quarrel\_polemicize}, {\sc reduce\_diminish}, {\sc refuse}, {\sc regret\_sorry}, {\sc remove\_take-away\_kidnap}, {\sc reprimand}, {\sc resign\_retire}, {\sc resist}, {\sc restrain}, {\sc slow-down}, {\sc spoil}, {\sc steal\_deprive}, {\sc stop}, {\sc subjugate}, {\sc violate}, {\sc waste}, {\sc weaken}, {\sc worsen}
\end{document}
%%% Local Variables:
%%% mode: LaTeX
%%% TeX-master: t
%%% End: